# BUY WHEN? SURVIVAL MACHINE LEARNING MODEL COMPARISON FOR PURCHASE TIMING.

Diego Vallarino[1]

[1]Independent Researcher, Madrid, Spain.
`diego.vallarino@gmail.com`

***ABSTRACT***

*The value of raw data is unlocked by converting it into information and knowledge that drives decision-making. Machine Learning (ML) algorithms are capable of analysing large datasets and making accurate predictions. Market segmentation, client lifetime value, and marketing techniques have all made use of machine learning.*

*This article examines marketing machine learning techniques such as Support Vector Machines, Genetic Algorithms, Deep Learning, and K-Means. ML is used to analyse consumer behaviour, propose items, and make other customer choices about whether or not to purchase a product or service, but it is seldom used to predict when a person will buy a product or a basket of products.*

*In this paper, the survival models Kernel SVM, DeepSurv, Survival Random Forest, and MTLR are examined to predict tine-purchase individual decisions. Gender, Income, Location, PurchaseHistory, OnlineBehavior, Interests, PromotionsDiscounts and CustomerExperience all have an influence on purchasing time, according to the analysis. The study shows that the DeepSurv model predicted purchase completion the best. These insights assist marketers in increasing conversion rates.*

***KEYWORDS***

*Survival Analysis, Machine Learning for survival, marketing analytics.*

# 1. INTRODUCTION

Due to advancements in information technology and the rapid rise of the Internet, the data revolution of the past several decades has caused businesses to create more data than they can utilize or understand (see Erevelles, S.; Fukawa, N.; Swayne, L., 2016; Seng, J.L.; Chen, T., 2010). The expansion in the volume of data, the variety of data kinds, and the scope of analysis has necessitated technological advancements beyond storage, transport, and processing (see Seng, J.L.; Chen, T., 2010)

The data must be translated into information and knowledge in order to transfer knowledge into decision-making tools for enterprises. This data is used in marketing research to identify intriguing links between market segmentation in industrial, tourist, and other markets, customer lifetime value, loyalty and client segment, direct market, marketing campaign, and other applications (Tkǎc, M.; Verner, R., 2016). With the application of Machine Learning (ML) methods (see Bahari, T.F.; Elayidom, M.S., 2015; Jessen, H.C.; Paliouras, G., 2001), it is currently anticipated that enormous volumes of stored data may be explored, and usable information extracted.

ML are strategies that equip computers with the capacity to comprehend, using data and experiences similar to the human brain (Çelik, Ö., 2018). The application of ML models to data is motivated by the fact that these models can be applied to the resolution of a variety of problems, ranging from those that could be solved through conventional statistics and management of scientific techniques to complex problems requiring a larger analysis; this is because ML allows solving problems more quickly and effectively than conventional methods. These strategies are used to forecast the outcome of new data, make predictions, classify information, or aid in the decision-making process.

Although many academics have worked with ML approaches in marketing, it is vital to summarize the numerous ways that may be used to certain areas, as well as how these methods might maximize outcomes and contribute to marketing-related firm decision making. Important to the study of the future of these techniques is the management of data; therefore, a review of the different methods commonly used in marketing has been conducted, with the goal of generating technologies and resolution of problems in a short period of time, with excellent results, such as Support Vector Machine (SVM), Genetic Algorithms (GA), Deep Learning, and K-Means, among others.

Several researchers have investigated marketing techniques. In the current state of the art, (see Sun, Z.; Guo, Q.; Yang, J.; Fang, H.; Guo, G.; Zhang, J.; Burke, R., 2019) describes the recommendation system's algorithms. They demonstrate the progression of approaches used to generate suggestions, such as collaborative algorithms for filtering based on proximity and model-based approximations.

The use of Artificial Neural Networks (ANN) for regression, analysis of primary components, and classification as a tool to estimate consumer behavior is described in (Briesch, R.; Rajagopal, P., 2018). They discuss the benefits and drawbacks of their usage of ANN in marketing and the application according to the evaluation of the disclosed by other authors in essence, behavior, satisfaction, customer segmentation, or the selection of a new product or response to publicity. (Chen, S.H.. 2016), illustrates the several methods that may be used to anticipate the departure of customers from various sorts of businesses (see Liu, X.; Singh, P.V.; Srinivasan, K., 2016).

These algorithms consist of, among others, support vector machines (SVM), decision trees (DT), and artificial neural networks (ANN). To conduct their study, authors such as (Balducci, B.; Marinova, D., 2018; Liu, X.; Singh, P.V.; Srinivasan, K., 2016) employ unstructured data. To estimate consumer demand for TV series, acquire data from internet sources such as Twitter Tweets, Google Trends, Wikipedia visits, IMDB ratings, and Huffington Post articles (see Liu,

X.; Singh, P.V.; Srinivasan, K., 2016). To analyze a large quantity of unstructured data, these writers use a variety of machine learning algorithms, text mining, and cloud computing, among others.

They observed that there is no clear correlation between the number of Twitter followers and the number of reviews acquired, projected the most-watched episodes, and concluded that this might have a future impact on demand, demographic distribution, and other elements that drive consumption. (Balducci, B.; Marinova, D., 2018), conduct a thorough analysis of the many approaches and software that may be used to handle unstructured data, and present a complete work framework to show the theoretical depth and computer advancements in marketing. Regarding customer decisions, (Hauser, J.R., 2014; Dzyabura, D.; Hauser, J.R., 2011) study the application of machine learning to the evaluation of the consumer's heuristics for selecting a series of items, employing a set of decision rules.

In Duarte, V., et al. (2018), it shows that marketing applications and ML approaches increased. Scientific productivity grew, but mainly in low-quality journals (Q2 and Q3) without the bottom quartile (Q4). Quotes decreased with time. ML approaches vary. In this era, deep learning was the most utilized approach, but in 2019, it became one of many. Due to greater technique-marketing subarea specificity, this is possible. They found that until 2008, marketing apps employed some tactics broadly, but in 2019, they became more targeted.

As a new approach, survival analyses focus on the time of the event of interest, complementing studies that examine whether a consumer will buy or act in a given manner. Survival analysis seldom predicts purchases. Time-varying variables and censoring in modeling make survival analysis better than static classification approaches.

The Cox proportional hazard model was initially used to predict bank failure for example, like Lane, Looney, and Wansley (1986) showed. Luoma and Laitinen (1991) used Cox proportional hazard models to predict Finnish industrial and retail firm failure, but they were inferior to discriminant and logit analysis. Shumway (2001) models discrete-time bankruptcy hazard using accounting and market data. Chava and Jarrow (2004), Carling, Pan, Ariyan, Narayan, and Truini (2007), Leong, Nguyen, Meredith, et al. (2008), and Leonardis & Rocci (2009) employed the discrete hazard model because of its parameter calculation advantages and the sort of variables organizations commonly submit (2008).

Logistic regression and Cox proportional hazards regression disregard repeating occurrences or fail to account for within-subject correlation, resulting in inaccurate standard error estimates and a departure from the original study issue (Twisk, Smidt, & de Vente, 2005). Many methods have been developed to analyze recurrent events using all available data and within-subject correlations. Marginal intensity approaches, based on various risk set definitions, allow all cases to be at risk for each repeated event (Wei, Lin, & Weissfeld, 1989), while conditional intensity models are estimated in elapsed time or gap time and only designate cases at risk for the kth repeated event after the (k-1)th event (Andersen & Gill, 1982; Chang & Wang, 1999; Prentice, Williams, & Peterson, 1981).

In the survey of the literature that we have carried out for this work, there is no relevant material that studies the time until purchase, using machine learning models under the survival methodology. As a consequence, we may conclude that survival analysis is not the major focus of specialists in purchase prediction. Our study seeks to evaluate the applicability of survival analysis (SA) to purchase decision.

In this study, we investigate the outcomes of our model comparison and their economic interpretation. Our investigation focuses on the ability of several models to forecast time-to-purchase based on a collection of pertinent data. Specifically, we examine the predictive power of the Kernel SVM, DeepSurv, Survival Random Forest, and MTLR survival models. We utilize the concordance index to compare the various machine learning techniques (C-index)

Our objective is to determine which model delivers the most accurate and useful forecasts of the time of purchase and to understand the economic relevance of the model's findings. To do this, we evaluate the relevance and size of the predicted coefficients for each model variable and compare these findings to economic theory and common sense. By studying the outcomes of our model comparison and the economic interpretation of these results, we seek to gain insight into the elements that lead to buy moment and a better knowledge of how various models may be used to anticipate these specific occurrences.

The remaining sections of the paper are organized as follows. Following is a description of the empirical analysis portion, including the used models, data source, and assessment measures. After presenting the results of the investigation, including a comparison of the various models, we examine the economic implications of our findings. Finally, we summarize the important results and their implications for future study and policymaking to end the work.

Overall, our work adds to the literature on the use of survival analysis in finance and sheds light on the causes that lead to financial collapses, therefore assisting policymakers in designing more effective rules to avoid similar disasters in the future.

## 2. EMPIRICAL ANALYSIS

In this section, we present the empirical analysis of the purchase moment. We analyze set of information from real-online customers, who have decided to buy or not to buy a specific basket of products, in a period of 12 months. The dataset consists of 10 variables, each of which contributes to understanding the factors influencing the timing of a purchase. The study of purchase moment is of great importance in marketing and behavioral economics, as it has significant implications retailer knowledge. In this section, we describe the models used in our analysis, the data source, and the evaluation metrics.

### 2.1 Models

#### 2.1.1 Cox Proportional Hazards Model (coxph)

The Cox proportional hazards model is a widely used semi-parametric model in survival analysis. It assumes that the hazard function can be represented as the product of a time-independent baseline hazard function and a time-varying covariate function. Mathematically, the model can be represented as:

$$h(t|x) = h_0(t) \exp(\beta^T x)$$

where $h(t|x)$ is the hazard function for a given time t and covariate values $x$, $h0(t)$ is the baseline hazard function, $\beta$ is a vector of regression coefficients, and $exp(\beta X)$ is the hazard ratio, which represents the change in hazard associated with a unit change in the covariate.

#### 3.1.2 Multi-Task Logistic Regression (MTLR)

Multi-task logistic regression is a machine learning method that can be used for survival analysis. It is a multi-output learning algorithm that can predict the probability of an event occurring at different time points. Mathematically, the model can be represented as:

$$h(t|x) = exp\left(\Sigma_{k=1}^{K}\Sigma_{j=1}^{p}\beta_{kj}x_{kj}\right)$$

Where $h(t|x)$ is the hazard rate for an individual with covariates x, $\beta_{kj}$ are the regression coefficients for the kth characteristic of the jth group, and $x_{kj}$ is the kth feature of the jth group.

### 2.1.3 Random Survival Forest

Random survival forests are an extension of random forests for survival analysis. They use an ensemble of decision trees to predict the survival function. The model can be represented as:

$$h(t|x) = (1/B)\Sigma_{b=1}^{B} h_b(t|x)$$

Where $h_b(t|x)$ is the hazard rate for an individual with covariates $x$ in the $bth$ decision tree and $B$ is the number of trees in the random forest.

### 2.1.4 DeepSurv

DeepSurv is a deep learning model for survival analysis. It uses a neural network with a flexible architecture to predict the survival function. The model can be represented as:

$$h(t|x) = exp\left(\Sigma_{i=1}^{p} \beta_i f_i(x) + g(h_\theta(x))\right)$$

Where $h(t|x)$ is the hazard rate for an individual with covariates $x$\$, \$$\beta\_i$ are the regression coefficients for the input features $f_i(x), g(\cdot)$ is a non-linear function that transforms the output features and $h_\theta(x)$ is a neural network with θ parameters.

## 2.2 Data

The dataset with which we have worked is a set of information from real-online customers, who have decided to buy or not to buy a specific basket of products, in a period of 12 months. The dataset consists of 1000 different people, with 10 variables, each of which contributes to understanding the factors influencing the timing of a purchase. Forty-eight percent of data censored by the right, which implies that the rest of the customers did buy the basket of products that was the focus of the investigation.

It contains information necessary for conducting an analysis on the time until a purchase of a specific basket of products . In this analysis, we will explore the components of each variable and examine their potential impact on the time to purchase.

### 2.2.1 Variable Components

*Age*: This variable represents the age of the individuals. Age can be a significant factor in determining purchasing behavior, as consumer preferences and needs often vary across different age groups. *Gender*: The gender variable indicates the gender of the individuals. Gender can play a role in shaping consumer preferences and purchase decisions, as it may influence product preferences, brand loyalty, and shopping behavior.

*Income*: The income variable reflects the income levels of the individuals. Income is an essential determinant of purchasing power and can influence the affordability and willingness to make a purchase. *Marital Status*: This variable captures the marital status of the individuals. Marital status can affect purchasing decisions, as the needs, priorities, and spending patterns of individuals may differ based on their marital status.

*Location*: The location variable represents the geographic location of the individuals. Geographical factors such as culture, availability of products, and local market dynamics can influence purchasing behavior. *Purchase History*: This variable indicates the individual's past

purchase history. Previous purchasing behavior can be an indicator of future purchase intentions, as individuals with a higher purchase history may exhibit greater brand loyalty or a higher propensity to make repeat purchases.

*Online Behavior*: The online behavior variable reflects the level of engagement of individuals in online activities related to shopping. Online behavior can influence the time to purchase, as individuals who are more active online may have access to a wider range of products, information, and promotions. *Interests*: This variable captures the specific interests of the individuals. Individual interests can influence purchase decisions, as individuals are more likely to purchase products or services related to their specific areas of interest.

*Promotions and Discounts*: The promotions and discounts variable indicate whether individuals are exposed to promotional offers and discounts. Promotional activities can influence the time to purchase by creating incentives for individuals to make a purchase sooner or by influencing their decision-making process. *Customer Experience*: This variable represents the individuals' past experiences with customer service. Customer experience can impact the time to purchase, as positive experiences may enhance brand perception and increase the likelihood of repeat purchases.

By analyzing these variables and their components, we aim to gain insights into the factors that contribute to the timing of a purchase. Understanding these factors can assist businesses in developing targeted marketing strategies, optimizing promotional activities, and improving customer satisfaction to enhance the overall sales performance and profitability.

## 2.3 Metrics

### 2.3.1 C-Index

The C-index (also known as the concordance index or the area under the receiver operating characteristic curve) is a widely used metric in survival analysis and medical research to assess the performance of predictive models that estimate the likelihood of an event occurring over a given time period.

The C-index is generated using the rankings of anticipated event occurrence probability for each participant in a dataset. It calculates the percentage of pairings of people in whom the person with the higher anticipated probability experienced the event before the person with the lower projected probability. In other words, it assesses a predictive model's capacity to rank people in order of their likelihood of experiencing the event of interest.

The C-index scales from 0 to 1, with 0.5 representing random prediction and 1 indicating perfect prediction. In medical research, a C-index value of 0.7 or above is considered satisfactory performance for a prediction model. Here is the formula of non-censored data C-Index.

$$C - index = \frac{\Sigma_{ij} 1_{T_j < T_i} \cdot 1_{\eta_j > \eta_i} \cdot \delta_j}{\Sigma_{ij} 1_{T_j < T_i} \cdot \delta_j}$$

$\eta_i$, the risk score of a unit $i$
$1_{T_j < T_i} = 0 \quad if \ T_j < T_i \ else \ 0$
$1_{\eta_j < \eta_i} = 0 \quad if \ \eta_j < \eta_i \ else \ 0$
$\delta_j$, represents whether the value is censored or not

# 3. RESULTS

The Kaplan-Meier curve presented below depicts the survival probability of time to purchase for a cohort of online buyers. In this context, the x-axis represents the duration of time, while the y-axis signifies the probability of surviving or completing a purchase. At the beginning of the observation period, all buyers are considered to have a survival probability of 1, indicating that they are all active participants.

As time progresses, some buyers may not complete their purchases within a specific time frame. In the context of this analysis, we refer to these incomplete purchases as "deaths."[1] Consequently, the survival probability gradually decreases over time, reflecting the diminishing likelihood of completing a purchase within the given timeframe.

From a microeconomic perspective, the decreasing survival probability for time to purchase suggests various factors at play. These may include consumer preferences, market conditions, competitive influences, or external events that impact the decision-making process of buyers. Microeconomists would delve deeper into these factors to gain insights into the underlying dynamics affecting the time to purchase.

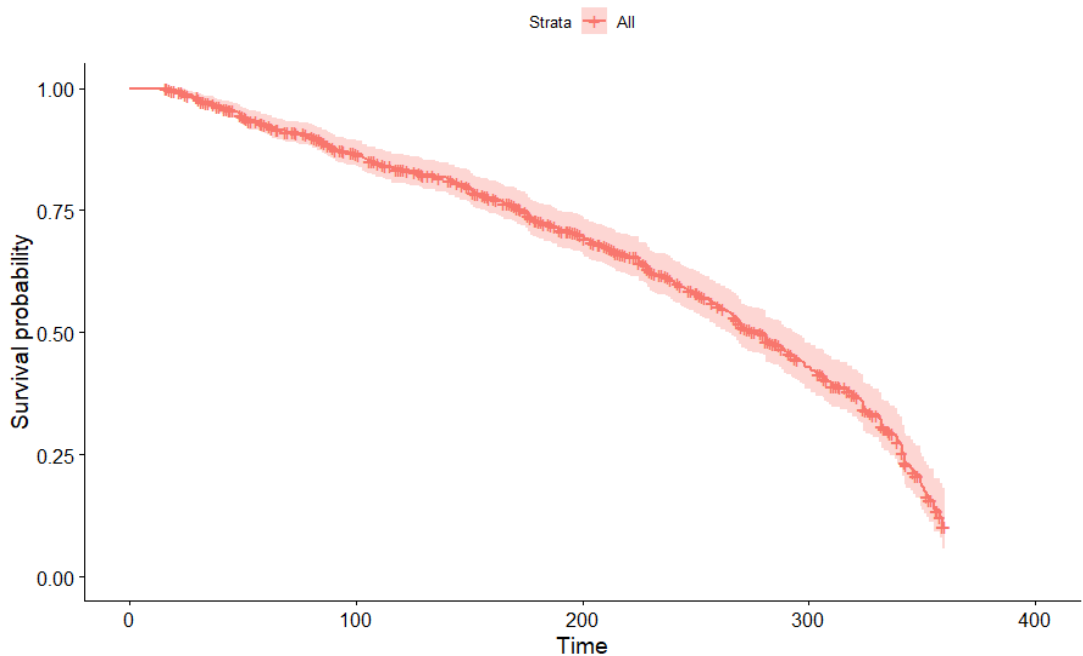

Figure 1: Kaplan-Meier survival curve.

---

[1] In survival analysis, the term "events" or "deaths" is commonly used to refer to the occurrence of the event of interest, which in this case would be completing a purchase. The survival probability, which represents the likelihood of surviving (or completing a purchase) beyond a certain time point, naturally decreases over time as more events or incomplete purchases occur.

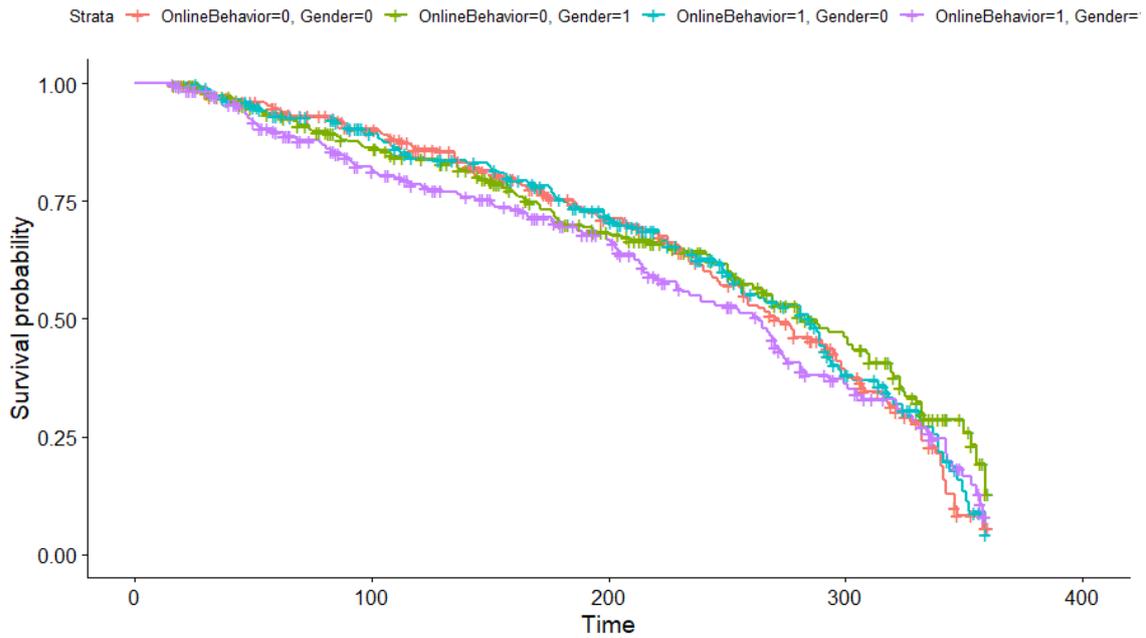

Figure 2: Kaplan-Meier survival curve considering online behavior and gender.

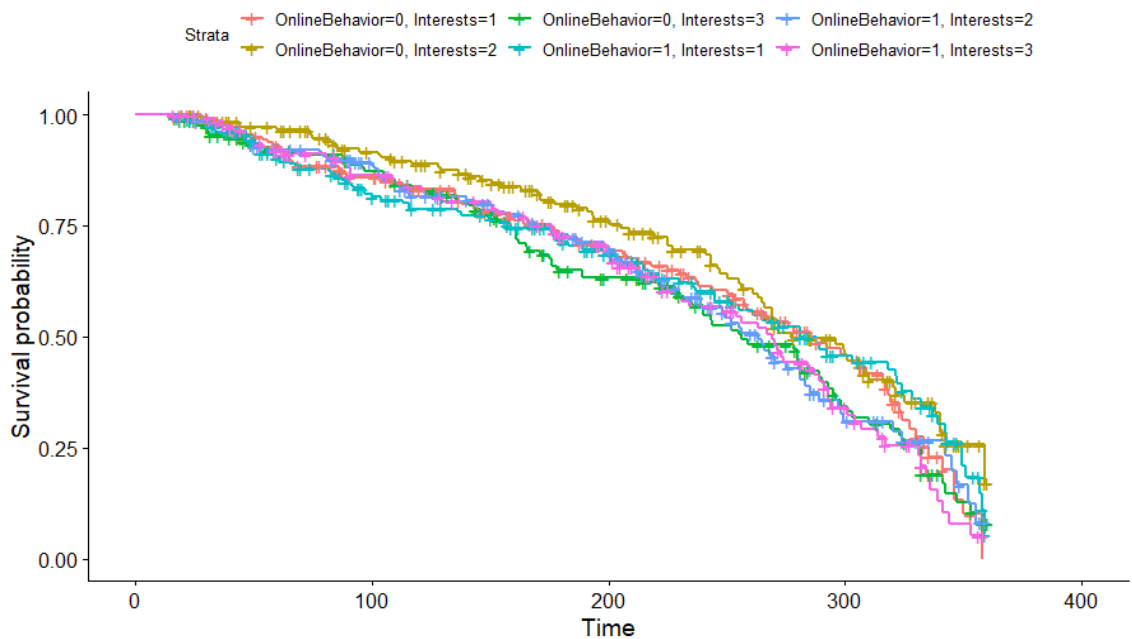

Figure 3: Kaplan-Meier survival curve considering online behavior and interests.

The figure above illustrates the changes in the risk of time to purchase over a given period. Initially, there were 395 instances of purchases at time 1.07, with one purchase occurring. This implies a high survival probability of 0.99747, indicating that most purchases were successfully completed. However, as time progressed to 4.07, the survival probability slightly decreased to 0.99494, indicating two additional instances of time to purchase.

From a microeconomic perspective, this decreasing survival probability suggests an increase in the risk of time to purchase over time. As we reach time 42.37, there were 382 instances remaining, but 13 purchases had occurred during the observation period. Consequently, the survival probability declined further to 0.96456, signifying a higher risk of delayed purchases.

Beyond time 42.37, the survival probability continued to rapidly decrease, suggesting a significant increase in the risk of time to purchase. This can be attributed to various factors such as changes in consumer behavior, market conditions, or competitive forces. A microeconomic analysis would delve deeper into these factors to understand the underlying causes of the increasing risk.

Analyzing the figure above, we can observe how the survival probability decreases as time progresses. For instance, at time 32, there were 664 instances at risk, with 18 events (time to purchase). The survival probability stood at 0.974, indicating a relatively low risk during this period. However, as time advanced to 350, only 23 instances remained, and 51 purchases occurred, resulting in a survival probability of 0.183. This indicates a significantly higher risk of delayed purchases at that point.

By examining these trends and patterns, marketers can gain valuable insights into the changing risk of time to purchase. This understanding allows them to adapt their strategies, refine targeting approaches, and optimize marketing efforts to mitigate the risk and improve conversion rates over time.

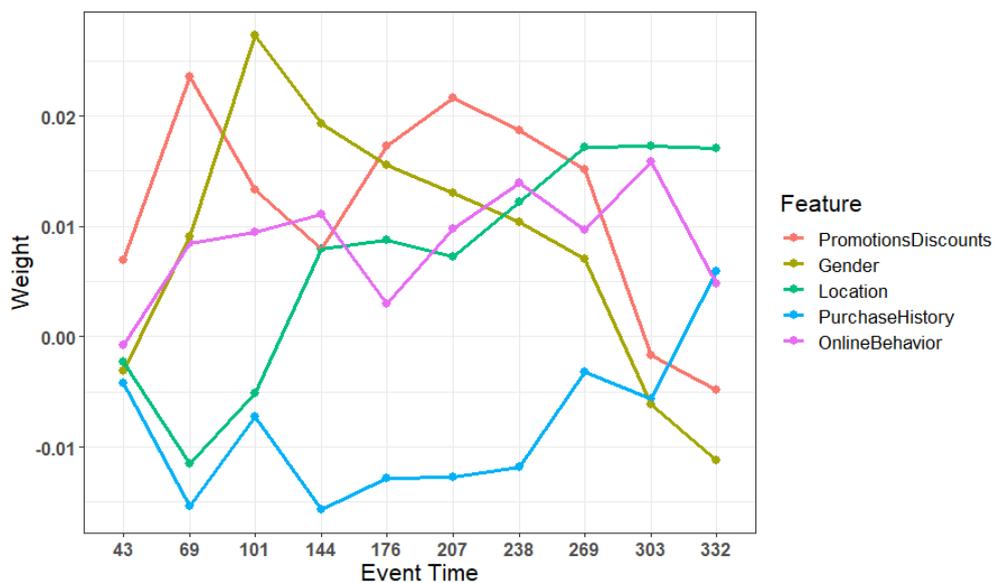

Figure 4: Relative weight of each variable based on MTLR model.

In the figure above, the variables that have the most weight to define the moment of a purchase are:

*PurchaseHistory:* The weight associated with the customer's purchase history has relatively larger values compared to other variables. It suggests that a customer's past purchase behavior is a strong predictor of the time to buy. Positive weights indicate that a higher purchase history is associated with a longer time to buy.

*PromotionsDiscounts:* The weight for promotions and discounts also has notable values. It indicates that offering promotions and discounts can influence the time to buy. Positive weights suggest that promotions and discounts are associated with a longer time to buy.

*CustomerExperience:* The weight for customer experience has a noticeable impact. It suggests that positive customer experiences can affect the time to buy. Positive weights indicate that a positive customer experience is associated with a longer time to buy.

*Age:* The age weight, although smaller in magnitude compared to the previous variables, still plays a role in predicting the time to buy. Positive weights imply that higher age values are associated with a longer time to buy.

*Gender, Income, MaritalStatus, Location, OnlineBehavior, and Interests:* These variables have smaller weights compared to the principal variables mentioned above. While they still contribute to the prediction, their impact might be relatively less significant. However, it is important to note that the relative importance of these variables may vary in different contexts.

### 3.1 What about *OnlineBehavior* and *Interests*?

When considering the time to buy a product basket, the weights associated with "*OnlineBehavior*" and "*Interests*" in the analysis output can provide insights into the factors that influence the duration it takes for customers to make a purchase. Here's how these variables can be interpreted in relation to the time to buy:

*OnlineBehavior*: the weights associated with "*OnlineBehavior*" indicate the impact of various online behaviors on the time to buy a product basket. Positive weights (e.g., 0.009428, 0.011134, 0.015881) suggest that certain online behaviors are associated with a longer time to buy. This implies that customers who exhibit these specific online behaviors might spend more time researching, comparing options, or engaging with digital channels before making a purchase decision.

On the other hand, negative weights (e.g., -0.00424, -0.01534, -0.01564) indicate that certain online behaviors are associated with a shorter time to buy. This suggests that customers displaying these behaviors might have a faster decision-making process, possibly due to higher confidence or familiarity with the product or brand. Analyzing specific online behaviors included in this variable can provide deeper insights into customers' digital engagement patterns and their impact on the time to purchase a product basket.

*Interests*: the weights associated with "Interests" offer insights into how customers' specific interests or preferences influence the time it takes for them to buy a product basket. Positive weights (e.g., 0.00294, 0.00475, 0.01321) indicate that certain interests are associated with a longer time to buy. This suggests that customers who have these specific interests might require more time to make a purchase decision, potentially due to the need for extensive research or consideration. Negative weights (e.g., -0.0131, -0.00931) suggest that certain interests are associated with a shorter time to buy.

This implies that customers with these interests might have a quicker decision-making process, possibly due to strong preferences or immediate needs related to the product basket. Analyzing the specific interests captured in this variable can help identify customer segments and tailor marketing strategies to align with their preferences, potentially influencing their time to buy.

By considering both *OnlineBehavior* and *Interests*, marketers can gain a more comprehensive understanding of the factors that contribute to the time it takes for customers to purchase a product basket. Leveraging these insights, marketers can develop targeted marketing campaigns, personalized recommendations, and customer experiences that align with customers' online behaviors and interests, ultimately optimizing the timing and relevance of their efforts to drive conversions.

## 3.2 Model comparison

The paper analyzed the performance of different machine learning survival models in predicting time-to-purchase using a set of relevant variables. The concordance index (C-index) was used to compare the predictive power of different models.

Based on the findings presented in the study, the performance of various survival models in predicting time to purchase was assessed using the C-index metric. Among the models considered, the DeepSurv model demonstrated the highest C-index value of 0.888524, indicating superior predictive accuracy compared to the other models. The RandForest model achieved the second-highest C-index of 0.724273, followed by the MTLR model with a C-index of 0.531151. The Cox model yielded a C-index of 0.507152, while the KarnelSVM model performed least effectively, obtaining the lowest C-index of 0.571.

These results suggest that the DeepSurv model outperformed the other models in predicting time to purchase. The RandForest and MTLR models also exhibited reasonably good predictive performance, although not as high as the DeepSurv model. This information holds significance for microeconomists, enabling them to understand and forecast the likelihood of purchase completion, aiding in decision-making processes and risk mitigation strategies.

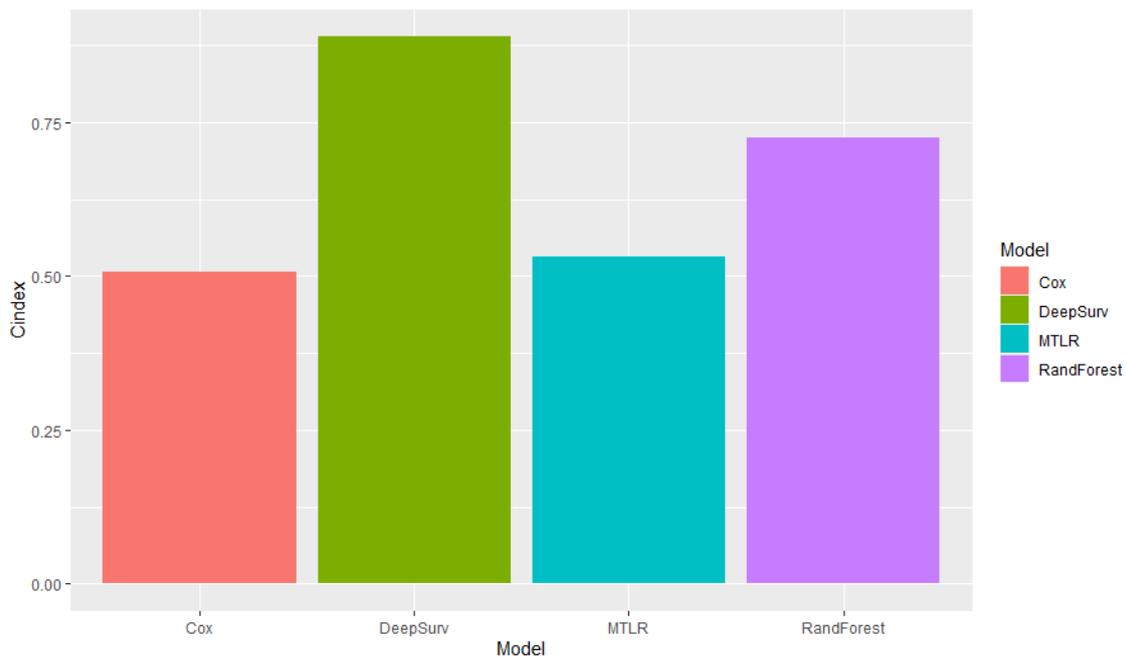

Figure 5: results from different machine learning models.

The variables examined in this study, such as Gender, Income, Location, PurchaseHistory, OnlineBehavior, Interests, PromotionsDiscounts and CustomerExperience, offer valuable insights into the factors contributing to understand what the moment of purchase will be, and its associated probability.

By comprehending the impact of these variables, stakeholders in the field of survival analysis can take appropriate measures to reduce the likelihood of incomplete purchases, leading to more accurate predictions and informed decision-making. As such, this research contributes to advancing the field of survival analysis in the domain of marketing and microeconomics.

# 4. CONCLUSION

The variables analyzed in this study provide valuable insights into the factors that influence the time to purchase. Purchase history emerges as the most significant variable, indicating that a customer's past purchase behavior strongly predicts the time it takes to make a purchase. Promotions and discounts, along with positive customer experiences, also play notable roles in influencing the duration. Age, although less impactful than the primary variables, still contributes to the prediction.

Regarding the variables "*OnlineBehavior*" and "*Interests*," they offer further understanding of customer behavior and preferences in relation to the time to buy. Certain online behaviors are associated with longer decision-making processes, while others are linked to quicker purchase decisions. Similarly, specific interests or preferences can either extend or shorten the time it takes to make a purchase.

By considering both online behavior and interests, marketers can gain comprehensive insights into customer segments and tailor their strategies accordingly. This knowledge can be leveraged to develop targeted marketing campaigns, personalized recommendations, and customer experiences aligned with customers' online behaviors and interests, thereby optimizing conversion rates.

Furthermore, the study demonstrates the effectiveness of machine learning survival models in predicting the time to purchase. The DeepSurv model exhibited superior predictive accuracy, followed by the RandForest and MTLR models. These models provide valuable tools for microeconomists to forecast purchase completion and make informed decisions to mitigate risks.

Based on our results, the research proposes a number of possible next actions and areas for development in order to increase our knowledge of the variables impacting the time-to-buy decisions. Analyzing a bigger and more varied sample may give a wider viewpoint and boost the results' generalizability. We may assess whether the observed associations remain true across other settings by including data from numerous firms or sectors. While the present research focused on a single organization, data from other companies or sectors would be useful for comparing the outcomes. This would aid in identifying similar trends and determining if the variables impacting purchase timing are constant across diverse situations.

Although the research looked at numerous major characteristics, there may be additional factors that influence the timing to buy. Additional factors might include macroeconomic situation, geographic region, or client preferences for certain product qualities. Extending the number of variables examined may offer a more complete picture of the dynamics at work. Other customer-related criteria, such as customer happiness, loyalty, or engagement, may give further insights into their influence on the time to buy in addition to the variables examined. Incorporating these variables may aid in determining whether they interact with or mediate the correlations reported in the present investigation.

Validating the results using independent external datasets would be good. By comparing the outputs across multiple datasets and establishing the stability of the correlations, this may assist assure the robustness and dependability of the results. Combining quantitative and qualitative research approaches, like as interviews or focus groups, may give a more in-depth insight of consumers' underlying motives and decision-making processes. This qualitative data may help to expand the understanding of quantitative findings and provide context for the results.

It is vital to emphasize the following restrictions that we examine at this time. The research was done on a single firm, limiting the results' generalizability to other companies or sectors. It is critical to recognize this restriction and evaluate the necessity for replication studies in various situations. It is critical that the data utilized in the analysis be accurate and full. Assessing the data's dependability and dealing with any possible data quality concerns, such as missing or skewed data, should be a top focus.

The study intended to discover links between attributes and the moment of purchasing. Creating causal links, on the other hand, requires more research, such as experimental designs or quasi-experimental approaches. The study may not have considered all external factors that may influence when to purchase, such as macroeconomic conditions, industry developments, or competitors' strategies. Taking these external aspects into consideration in future studies may provide a fuller understanding of the dynamics involved.

Researchers may continue to progress the discipline of survival analysis and increase our knowledge of the variables influencing the time to buy by addressing these next steps and constraints, allowing companies to make more informed choices and improve their marketing tactics.

**Diego Vallarino, PhD**

Short Biography

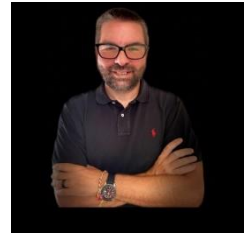

I am an experienced C-level executive and Principal Data Scientist with over 20 years of experience in the financial services and information business industries. My expertise lies in business development using data science. I have held prominent positions such as Global Head Data Discovery at Coface Group, Chief Data Scientist at Scotiabank, Data Analytics Director at Equifax, and Senior International Consultant for organizations like the World Bank, IADB, and UNDP.

In 2023, I was awarded the Einstein Visa (EB1A Green Card) by the US government. I have been recognized as a "Global Top 100 Data Visionary" by Truata in 2020 and received the Global One Equifax Award in 2018. Currently, I also serve as a Board Member for UTEC University in Peru and am a full member of CDO LatAm (Peru) since 2020 and LatinxInAI (San Francisco, US) since 2022. Additionally, I am a member of The Economic History Society since 2021 and the South American Network of Applied Economics since 2009.

As an author, I published the book "Innovation from the South" in 2005, and I have delivered over 40 international conferences worldwide.

My academic background includes a PhD in Cliometrics from Universidad Torcuato Di Tella (Argentina) completed between 2016 and 2018. I also hold an MSc in (bio)Statistics from the University of Barcelona (Spain) earned between 2020 and 2022, and an MSc in Data Management & Big Data from the same university in 2015. I furthered my education with an MBA Full Time from Universidad Adolfo Ibáñez (Chile) in 2002-2003 and obtained a BSc in Business & CPA from FCEA - UdelaR (Uruguay) between 1994 and 2000.

Throughout my career, I have pursued continuous learning through post-academic degrees and executive education courses. Notable among them are SEE from Babson College (USA, 2007), GCP from The Wharton School (USA, 2003), MOC in Microeconomics of Competitiveness from Harvard Business School (USA, 2010), and Data Science Ethics from the University of Michigan (USA, 2018).